\documentclass{article}

     \usepackage[preprint]{neurips_2019}

\usepackage[utf8]{inputenc} % allow utf-8 input
\usepackage[T1]{fontenc}    % use 8-bit T1 fonts
\usepackage{hyperref}       % hyperlinks
\usepackage{url}            % simple URL typesetting
\usepackage{booktabs}       % professional-quality tables
\usepackage{amsfonts}       % blackboard math symbols
\usepackage{nicefrac}       % compact symbols for 1/2, etc.
\usepackage{microtype}      % microtypography
\usepackage{graphicx}
\usepackage{amsmath}
\usepackage{amssymb}
\usepackage{pifont}
\usepackage{makecell}
\usepackage{subcaption}
\usepackage{hyperref}
\usepackage{url}
\usepackage{xcolor}

\title{One-Shot Weakly Supervised Video Object Segmentation}

\author{ Mennatullah Siam\thanks{equally contributing}\\
University of Alberta\\
{\tt\small mennatul@ualberta.ca}
\And
Naren Doraiswamy\textsuperscript{*}\\
Indian Institute of Science (IISc)\\
{\tt\small narend@iisc.ac.in}
\And
Boris N. Oreshkin\textsuperscript{*}\\
ElementAI\\
{\tt\small boris@elementai.com}
\And
Hengshuai Yao\\
HiSilicon, Huawei Research\\
{\tt\small hengshuai.yao@huawei.com}
\And
Martin Jagersand\\
University of Alberta\\
{\tt\small jag@cs.ualberta.ca}
}

\begin{document}

\maketitle

\begin{abstract}
Conventional few-shot object segmentation methods learn object segmentation from a few labelled support images with strongly labelled segmentation masks. Recent work has shown to perform on par with weaker levels of supervision in terms of scribbles and bounding boxes. However, there has been limited attention given to the problem of few-shot object segmentation with image-level supervision. We propose a novel multi-modal interaction module for few-shot object segmentation that utilizes a co-attention mechanism using both visual and word embeddings. It enables our model to achieve 5.1\% improvement over previously proposed image-level few-shot object segmentation. Our method compares relatively close to the state of the art methods that use strong supervision, while ours use the least possible supervision. We further propose a novel setup for few-shot weakly supervised video object segmentation(VOS) that relies on image-level labels for the first frame. The proposed setup uses weak annotation unlike semi-supervised VOS setting that utilizes strongly labelled segmentation masks. The setup evaluates the effectiveness of generalizing to novel classes in the VOS setting. The setup splits the VOS data into multiple folds with different categories per fold. It provides a potential setup to evaluate how few-shot object segmentation methods can benefit from additional object poses, or object interactions that is not available in static frames as in PASCAL-$5^i$ benchmark.
\end{abstract}

\section{Introduction}
The few-shot learning literature has mainly focused on the classification task such as~\cite{koch2015siamese}\cite{vinyals2016matching}\cite{snell2017prototypical}\cite{qi2017learning}\cite{finn2017model}\cite{ravi2016optimization}\cite{sung2018learning}\cite{qiao2018few}. Recently, solutions for few-shot object segmentation which learns pixel-wise classification based on few labelled samples have emerged~\cite{shaban2017one}\cite{rakelly2018conditional}\cite{dong2018few}\cite{wang2019panet}\cite{zhang2019pyramid}\cite{zhang2019canet}\cite{siam2019amp}. Previous literature in few-shot object segmentation relied on manually labelled segmentation masks. A few recent works~\cite{rakelly2018conditional}\cite{zhang2019canet}\cite{wang2019panet} started to conduct experiments using weak annotations in terms of scribbles, bounding boxes. However, limited research was conducted on using image-level supervision for few-shot object segmentation with one sole recent work~\cite{raza2019weakly}.

In order to improve image-level few-shot object segmentation we propose a multi-modal interaction module that leverages the interaction between support and query visual features and word embeddings. The use of neural word embeddings pretrained on GoogleNews~\cite{mikolov2013distributed} and visual embeddings from pretrained networks on ImageNet~\cite{deng2009imagenet} allows to build such a model. We propose a novel approach that uses Stacked Co-Attention to leverage the interaction among visual and word embeddings. We mainly inspire from~\cite{lu2019see} which proposed a co-attention siamese network for unsupervised video object segmentation. However, our setup mainly focuses on the few-shot object segmentation aspect to assess its ability to generalize to novel classes. This aspect motivates why we meta-learned a multi-modal interaction module that leverages the interaction between support and query with image-level supervision.

Concurrent to few-shot object segmentation, video object segmentation(VOS) has been extensively researched~\cite{Pont-Tuset_arXiv_2017}\cite{Perazzi2016}\cite{lu2019see}\cite{tokmakov2016learning}\cite{jain2017fusionseg}\cite{luiten2018premvos}. Two main categories for video object segmentation are studied which are semi-supervised VOS and unsupervised VOS. The semi-supervised VOS also named as few-shot requires an initial strongly labelled segmentation mask to be provided. In a recent work~\cite{khoreva2018video} video object segmentation using language expression has been studied. However, their work does not focus on the aspect of segmenting novel categories which we try to address since it can be of potential benefit to the few-shot learning community. The additional temporal information in VOS can provide extra unlabelled data about the object poses or its interactions with other objects to the few-shot object segmentation method. It has the potential to be closer to human-like learning rather than learning from a single frame. To the best of our knowledge we are the first to propose a novel setup for video object segmentation that focuses on the ability to generalize to novel classes by splitting the categories to different folds which would better assess the generalization ability. The setup as well only requires image-level labels for the first frame to segment the corresponding objects unlike conventional semi-supervised VOS. Although Youtube-VOS~\cite{xu2018youtube} provided a way to evaluate on certain unseen categories but it does not utilize any of the category label information in the segmentation model. However, in order to ensure the evaluation for the few-shot method is not biased to a certain category, it is best to split into multiple folds and evaluate on different ones. To this end we use Youtube-VOS dataset~\cite{xu2018youtube} which is a large-scale VOS dataset with 65 different object categories, the originally provided training data is further split into 5 different folds with 13 novel classes per fold. The novel setup opens the door towards studying how few-shot object segmentation can benefit from temporal information for the different object poses or its interactions. The main contributions in this paper are summarized as follows: 
\begin{itemize}
    \item A novel formulation for learning image-level labelled few-shot segmentation based on a multi-modal interaction module is presented. It utilizes neural word embeddings and attention mechanisms that relates the support and query images. It enables attention to the most relevant spatial locations in the query feature maps.
    \item We propose a novel setup called few-shot weakly supervised video object segmentation that can be of potential benefit to few-shot object segmentation.
\end{itemize}

\section{Method}
 \subsection{Stacked Co-Attention}
 \label{sec:CS2QA}
 A simple conditioning on the support set image would be insufficient where neither sparse nor dense annotations is provided. In order to leverage the interaction between the support set images and query set image a conditioned support-to-query attention module is used to learn the correlation between them. Initially a base network is used to extract features from support set image $I_s$ and query image $I_q$ which we denote as $V_s \in R^{W \times H \times C}$ and $V_q \in R^{W \times H \times C}$. Where $H$ and $W$ denote the feature maps height and width respectively, while $C$ denote the feature channels.

 It is important to initially capture the class label semantic representation through using semantic word embeddings~\cite{mikolov2013distributed} before performing the co-attention. The main reason is that both the support set and query set can contain multiple common objects from different classes, so depending solely on the support-to-query attention will fail in this case. A projection layer is used on the semantic word embeddings to construct $z \in R^{d}$ where $d$ is 256. It is then spatially tiled and concatenated with the visual features resulting in $\tilde{V_s}$ and $\tilde{V_q}$. An affinity matrix $S$ is then computed to capture the similarity between them using equation \ref{eq:coatt}.
\begin{equation}
    S = \tilde{V_s}^TW_{co}\tilde{V_q}
    \label{eq:coatt}
\end{equation}

\begin{figure*}[t]
\centering
    \includegraphics[width=0.7\textwidth]{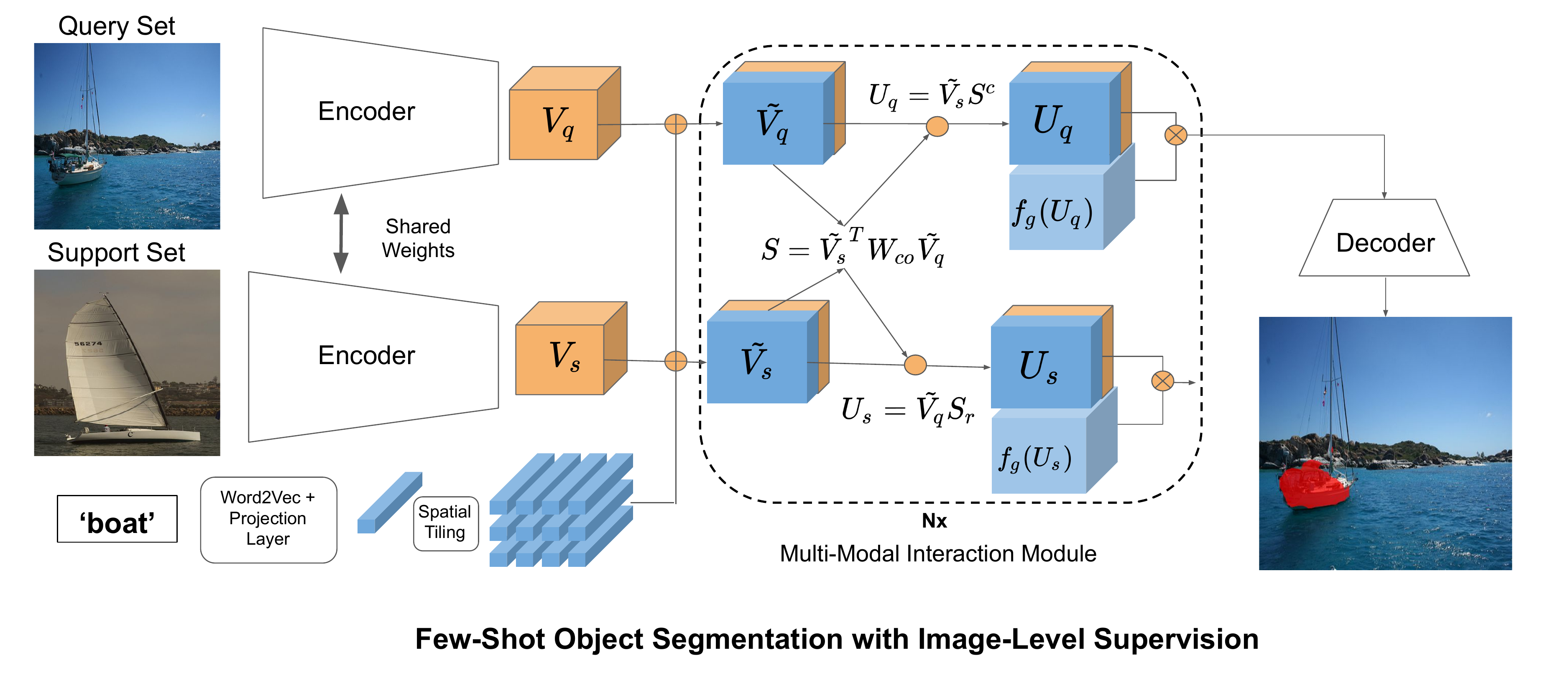}
    \caption{Architecture of Few-Shot Object segmentation model with co-attention. The $\oplus$ operator denotes concatenation, $\otimes$ denotes element-wise multiplication, $\circ$ denotes matrix multiplication.}
     \label{fig:detailed}
\end{figure*}

The feature maps are flattened into matrix representations where $\tilde{V_q} \in R^{C \times WH}$ and $\tilde{V_s} \in R^{C \times WH}$, while $W_{co} \in R^{C \times C}$ learns the correlation between feature channels. We use a vanilla co-attention similar to~\cite{lu2019see} where $W_{co}$ is learned using a fully connected layer. The resulting affinity matrix $S \in R^{WH \times WH}$ relates each column from $\tilde{V_q}$ and $\tilde{V_s}$. A softmax operation is performed on the $S$ row-wise and column-wise depending on what relation direction we are interested in following equations~\ref{eq:softmaxc} and \ref{eq:softmaxr}.
 \begin{subequations} 
 \begin{equation}
 S^c = \textup{softmax}(S) 
 \label{eq:softmaxc}
 \end{equation}
  \begin{equation}
 S^r = \textup{softmax}(S^T) 
 \label{eq:softmaxr}
 \end{equation}
 \end{subequations}
 $S^c_{*,j}$ has the relevance of the $j^{th}$ spatial location in $V_q$ with all spatial locations of $V_s$, where $j=1,...,WH$. The normalized affinity matrix can be used to compute $U_q$ using equation~\ref{eq:att_summs} and $U_s$ similarly. $U_q$ and $U_s$ act as the attention summaries.
  \begin{equation}
 U_q = \tilde{V_s} S^c
 \label{eq:att_summs}
 \end{equation}
 The attention summaries are further gated using a gating function $f_g$ with learnable weights $W_g$ and bias $b_g$ following equations~\ref{eq:gate1} and~\ref{eq:gate2}. The gating function ensures the output to be in the interval [0, 1] in order to mask the attention summaries using a sigmoid activation function $\sigma$. The $\circ$ operator denotes the hadamard product or element-wise multiplication.
 \begin{subequations}
 \begin{equation}
 f_g(U_q) = \sigma{(W_g U_q + b_g)} \in [0, 1]
 \label{eq:gate1}
 \end{equation}
 \begin{equation}
 U_q = f_g(U_q) \circ U_q
 \label{eq:gate2}
 \end{equation}
 \end{subequations}
 The gated attention summaries $U_q$ are concatenated with the original visual features $V_q$ and reshaped back to construct the final output from the attention module. Figure~\ref{fig:detailed} demonstrates our proposed method. We utilize a ResNet-50~\cite{he2016deep} encoder pre-trained on ImageNet~\cite{deng2009imagenet} to extract visual features. The segmentation decoder is comprised of an iterative optimization module (IOM)~\cite{zhang2019canet} and an atrous spatial pyramid pooling (ASPP)~\cite{chen2017deeplab}\cite{chen2017rethinking}. In order to improve our model we use two stacked co-attention modules. It allows the model to learn a better representation by letting the support set guide the attention on the query image and the reverse with respect to the support set through multiple iterations.
 
\subsection{Few-Shot Weakly Supervised Video Object Segmentation}
We propose a novel setup for few-shot video object segmentation where we utilize image-level labelled first frame to learn object segmentation in the sequence rather than using manual segmentation masks. More importantly the setup is devised in a way to split the categories to multiple folds to assess the generalization ability to novel classes. We utilize Youtube-VOS dataset training data which has 65 categories, and we split into 5 folds. Each fold has 13 classes that are used as novel classes, while the rest are used in the meta-training phase. A randomly sampled category $Y_s$ and sequence $V = \{I_1, I_2, ..., I_N\}$ is used to construct support set $S_p=\{(I_1, Y_s)\}$ and query images $I_i$.

\section{Experimental Results}
In this section we demonstrate results from experiments conducted on the PASCAL-$5^i$ dataset~\cite{shaban2017one} which is the most widely used dataset for evaluating few-shot segmentation. We also conduct experiments on our novel few-shot weakly supervised video object segmentation.

\subsection{Experimental Setup}
 We report two evaluation metrics, the mean-IoU computes the intersection over union for all 5 classes within the fold and averages them neglecting the background~\cite{shaban2017one}. Whereas the binary-IoU metric proposed in~\cite{rakelly2018conditional}\cite{dong2018few} computes the mean of foreground and background IoU in a class agnostic manner. Both metrics are reported as an average of 5 different runs to ensure a stable result following~\cite{wang2019panet}. We have also noticed some deviation in the validation schemes used in previous works. ~\cite{zhang2019canet} follows a procedure where the validation is performed on the $L_{test}$ classes to save the best model whereas~\cite{wang2019panet} does not perform validation and rather trains for a fixed number of iterations. We choose the approach followed in~\cite{wang2019panet} since we consider that the $L_{test}$ classes are not available during the initial meta-training phase.

\subsection{Few-shot Weakly Supervised Object Segmentation}
\begin{figure*}[ht!]
\begin{subfigure}{.33\textwidth}
    \centering
    \begin{subfigure}{.46\textwidth}
        \includegraphics[scale=0.12]{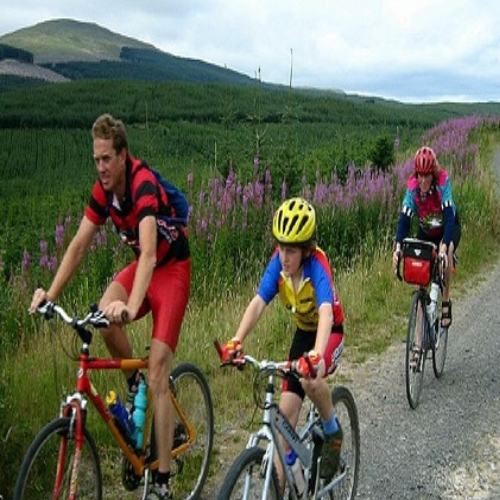}
    \end{subfigure}%
    \begin{subfigure}{.43\textwidth}
        \includegraphics[scale=0.12]{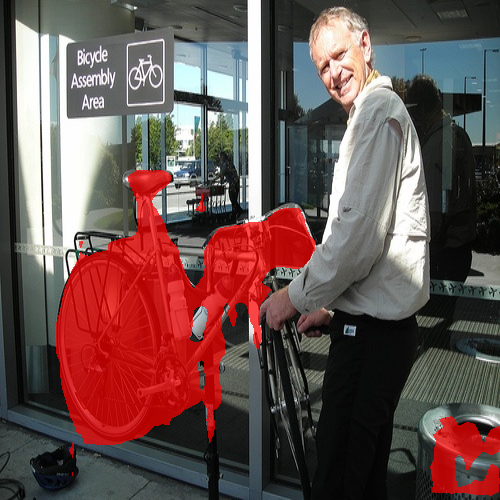}
    \end{subfigure}%
    \caption{'bicycle'}
\end{subfigure}%
\begin{subfigure}{.33\textwidth}
    \centering
    \begin{subfigure}{.46\textwidth}
        \includegraphics[scale=0.12]{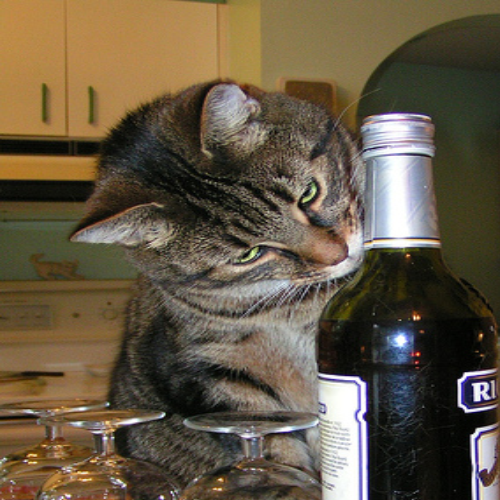}
    \end{subfigure}%
    \begin{subfigure}{.43\textwidth}
        \includegraphics[scale=0.12]{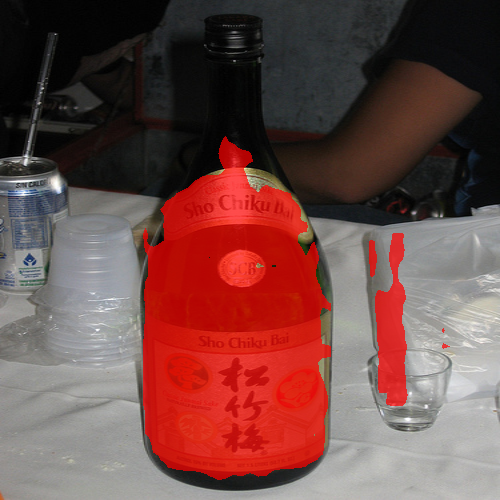}
    \end{subfigure}%
    \caption{'bottle'}
\end{subfigure}%
\begin{subfigure}{.33\textwidth}
    \centering
    \begin{subfigure}{.46\textwidth}
        \includegraphics[scale=0.12]{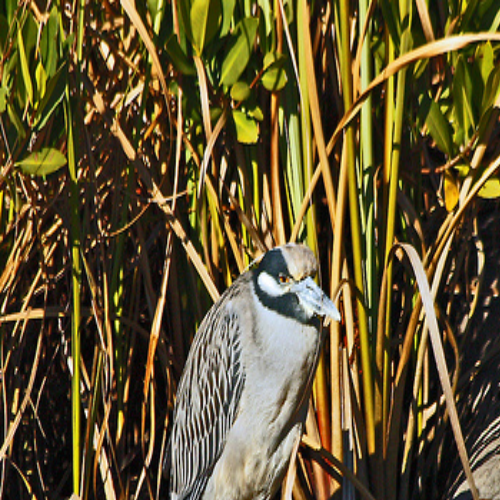}
    \end{subfigure}%
    \begin{subfigure}{.43\textwidth}
        \includegraphics[scale=0.12]{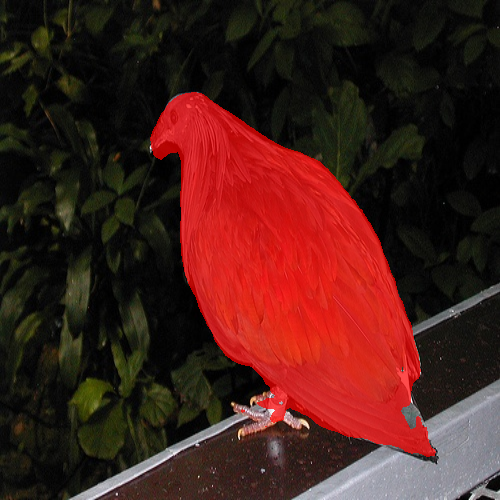}
    \end{subfigure}%
    \caption{'bird'}
\end{subfigure}

\caption{Qualitative evaluation on PASCAL-$5^i$ 1-way 1-shot. The support set and prediction on the query image are shown in pairs.}
\label{fig:pascal5i}
\end{figure*}

We compare the result of our proposed method with stacked co-attention against the other state of the art methods for 1-way 1-shot segmentation on pascal-$5^i$ in Table~\ref{table:pascal5i} using mean-IoU and binary-IoU metrics. We report the results for both the two validation schemes where V1 is following~\cite{zhang2019canet} and V2 is following~\cite{wang2019panet} validation scheme. Without the utilization of segmentation mask or even sparse annotations, our method with the least supervision of image level labels performs relatively on-par 53.5\% compared to the current state of the art methods 56\% showing the efficacy of our proposed algorithm. It outperforms the previous one-shot weakly supervised segmentation~\cite{raza2019weakly} with 5.1\%. Our proposed model outperforms the baseline method which utilizes a co-attention module without using word embeddings. Figure~\ref{fig:pascal5i} shows the qualitative results for our proposed approach.

\newcommand{\xmark}{\ding{55}}%
\newcommand{\cmark}{\ding{51}}
\begin{table}[t!]
\caption{Quantitative results for 1-way, 1-shot segmentation on the PASCAL-$5^i$ dataset showing mean-Iou~\cite{shaban2017one} and binary-IoU~\cite{rakelly2018conditional}\cite{dong2018few}. W: stands for using weak supervision from Image-Level labels.}
\centering
\begin{tabular}{|l|c|cccc|c|c|}
\hline
 Method & W & fold 0 & fold 1 & fold 2 & fold 3 & mean-IoU & binary-IoU\\ \hline
FG-BG & \xmark & - & - & - & - & - & 55.1 \\ 
\cite{shaban2017one} & \xmark & 33.6 & 55.3 & 40.9 & 33.5 & 40.8 & - \\
\cite{rakelly2018conditional} & \xmark & 36.7 & 50.6 & 44.9 & 32.4 & 41.1 & 60.1\\ 
\cite{dong2018few} & \xmark & - & - & - & - & - & 61.2 \\ 
\cite{siam2019amp} & \xmark & 41.9 & 50.2 & 46.7 & 34.7 & 43.4 & 62.2 \\ 
\cite{wang2019panet} & \xmark & 42.3 & 58.0 & 51.1 & 41.2 & 48.1 & 66.5 \\ 
\cite{zhang2019canet} & \xmark & 52.5 & 65.9 & 51.3 & 51.9 & 55.4 & 66.2\\ 
\cite{zhang2019pyramid} & \xmark & 56.0 & 66.9 & 50.6 & 50.4 & 56.0 & 69.9 \\ \hline
\cite{raza2019weakly} & \cmark & - & - & - & - & - & \color{blue}{\textbf{58.7}} \\
Baseline & \cmark & 38.6 & 56.6 & 43.8 & 38.2 & 44.3 & 60.2 \\
Ours-V1 & \cmark & 49.4 & 65.5 & 50.0 & 49.2 & 53.5 & \color{red}{\textbf{65.6}} \\
Ours-V2 & \cmark & 42.1 & 65.1 & 47.9 & 43.8 & 49.7 & \color{red}{\textbf{63.8}} \\ \hline
\end{tabular}
\label{table:pascal5i}
\end{table}

\subsection{Few-shot Weakly Supervised Video Object Segmentation}
Table~\ref{table:ytbvos} shows results on our proposed novel setup and comparing our method with the baseline of using co-attention module without utilizing word embeddings similar to~\cite{lu2019see}. It shows the potential benefit from utilizing neural word embeddings to guide the attention module.

\begin{table}[h]
\caption{Quantitative Results on Youtube-VOS One-shot weakly supervised setup.}
\centering
\begin{tabular}{|c|c|l|l|l|l|l|}
\hline
Method & fold 0 & fold 1 & fold 2 & fold 3 & fold 4 & Mean \\ \hline
Baseline & 40.1 & 33.7 & 47.1 & 36.4 & 36.6 & 38.8 \\ \hline
Ours & 41.6 & 40.8 & 51.4 & 41.5 & 39.1 & \textbf{42.9} \\ \hline
\end{tabular}
\label{table:ytbvos}
\end{table}

\section{Conclusions}
Our proposed method demonstrates great promise toward performing few-shot object segmentation based on gated co-attention that leverages the interaction between the support set and query image. Our model utilizes neural word embeddings to guide the attention mechanism which improves the segmentation accuracy compared to the baseline. We demonstrate promising results on Pascal-$5^i$ where we outperform the previously proposed image-level labelled one-shot segmentation method by 5.1\% and perform closer to methods that use strongly labelled masks. Our novel setup provides a mean to experiment with the effect of capturing different object viewpoints, and appearance changes in few-shot object segmentation. It closely mimics human learning for novel objects from few labelled samples by aggregating information from different viewpoints and capturing different object interactions.

\bibliography{neurips_2019}
\bibliographystyle{neurips_2019}

\end{document}